**Improving Crash Data Quality with Large Language Models: Evidence from Secondary Crash Narratives in Kentucky**

**Xu Zhang, Ph.D.**
(Corresponding Author)
Research Scientist
Kentucky Transportation Center,
266 Raymond Bldg, Lexington, KY 40506-0281
Phone: (859) 257-8037
Email: xuzhang_uk@uky.edu

**Mei Chen, Ph.D.**
Professor
Department of Civil Engineering, University of Kentucky
267 Raymond Bldg, Lexington, KY 40506-0281
Phone: (859) 257-9262
Email: mei.chen@uky.edu

**ABSTRACT**

High-quality crash data is essential for effective traffic safety analysis, yet police-reported crash databases often suffer from underreporting and miscoding, particularly for secondary crashes. This study evaluates advanced natural language processing (NLP) techniques to enhance crash data quality by mining crash narratives, using secondary crash identification in Kentucky as a case study. Drawing from 16,656 manually reviewed narratives from 2015–2022, with 3,803 confirmed secondary crashes, we compare three model classes: zero-shot open-source large language models (LLMs) (LLaMA3:70B, DeepSeek-R1:70B, Qwen3:32B, Gemma3:27B); fine-tuned transformers (BERT, DistilBERT, RoBERTa, XLNet, Longformer); and traditional logistic regression as baseline. Models were calibrated on 2015–2021 data and tested on 1,771 narratives from 2022. Fine-tuned transformers achieved superior performance, with RoBERTa yielding the highest F1-score (0.90) and accuracy (95%). Zero-shot LLaMA3:70B reached a comparable F1 of 0.86 but required 139 minutes of inference; the logistic baseline lagged well behind (F1: 0.66). LLMs excelled in recall for some variants (e.g., GEMMA3:27B at 0.94) but incurred high computational costs (up to 723 minutes for DeepSeek-R1:70B), while fine-tuned models processed the test set in seconds after brief training. Further analysis indicated that mid-sized LLMs (e.g., DeepSeek-R1:32B) can rival larger counterparts in performance while reducing runtime, suggesting opportunities for optimized deployments. Results highlight trade-offs between accuracy, efficiency, and data requirements, with fine-tuned transformer models balancing precision and recall effectively on Kentucky data. Practical deployment considerations emphasize privacy-preserving local deployment, ensemble approaches for improved accuracy, and incremental processing for scalability, providing a replicable scheme for enhancing crash-data quality with advanced NLP.

# 1. INTRODUCTION

High-quality crash data forms the foundation of traffic safety analysis. State and local agencies rely on police crash databases to locate crash hotspots, reveal causal relationships, select countermeasures, and prioritize safety investments. However, it is known that these databases suffer from data quality issues, including underreporting of crashes and injuries and miscoding of key crash attributes (1, 2). For example, a recent systematic review of police–hospital data linkages (covering 1994–2023) found that police crash reports consistently under-count clinically significant injuries across many settings, with pedestrian and cyclist injuries often highly under-reported (3). Likewise, critical factors such as alcohol or drug impairment and cell phone use are frequently under-recorded in crash reports (4). Beyond missing cases, crash records are often affected by inaccuracies such as missing fields, typo errors, and misclassification of important variables (1, 2). For instance, a statewide analysis in Kentucky found that only 8-13% of the crashes coded as "secondary" were truly secondary crashes in 2015-2017, while many actual secondary crashes went unmarked (5). These data quality issues undermine the validity of safety performance measures that agencies use to allocate safety resources.

Crash narratives, written by police officers detailing crash circumstances, offer a valuable resource to cross-check and improve the coded crash data. A narrative can mention nuanced details, such as "traffic was moving very slow due to an injury accident ahead", "debris from the first collision struck unit 2", or "a construction truck lost traction and collided with guard rail". Such circumstantial contexts provide a second chance to review the coded fields and flag inconsistent records for investigation and correction. In current practice, traffic engineers read these reports manually, which is labor-intensive and inconsistent. In a recent multi-university collaboration to improve crash data quality through narrative review, researchers at the Kentucky Transportation Center developed a proprietary web-based quality control tool to allow reviewers more easily identify discrepancies between narratives and coded data (6). Seven students from three universities were trained to verify 20 coded crash attributes against each narrative. Due to the labor-intensive nature of the reviewing process (approximately 3 minutes per narrative), the team managed to examine only 8,000 crashes, leaving much of the crash database unchecked.

Early machine-learning studies attempted to scale narrative review using "bag-of-words" text classification models such as logistic regression and support-vector machines (7-9). While these conventional methods are transparent and easy to deploy, they are sensitive to sparse vocabularies and cannot capture linguistic context, often resulting in high false positives and false negatives in practice. For example, the logistic regression-based classifier misidentified roughly one in four secondary crashes(9). Such results underscore the shortcomings of basic text-mining approaches in handling the complexity and ambiguity of crash narratives.

Recent progress in natural language processing have opened new opportunities for crash narrative mining. Transformer-based large language models (LLMs) with billions of parameters pre-trained on massive text databases exhibit exceptional capabilities in understanding nuanced syntax and semantics. These capabilities enable efficiently processing of large volume of crash narratives to extract useful insights often overlooked by conventional word-frequency approaches (10). Indeed, early experiments in transportation safety indicate that LLMs can reason over complex crash descriptions, infer contributing factors, and even explain their decisions (11). For example, Mumtarin et al. showed that public LLMs like ChatGPT, Bard, and GPT-4 were adept at answering complex questions about crash scenarios that traditional machine learning models struggled with (12). However, this potential lacks systematic benchmarking against established methods for crash data quality enhancement, particularly amid rapid LLM evolution through 2025.

This study addresses this critical gap by providing the most extensive evaluation to date of narrative-mining algorithms for crash data quality. We focus on secondary crash identification as a challenging case study, given its importance in Traffic Incident Management, the difficulty in distinguishing them from secondary events within a single crash, and the complexity of varied causal factors. Various approaches have been proposed in the past to identify secondary crashes, including fixed spatiotemporal thresholds, shockwave theory methods, dynamic speed profile analyses, and text mining via traditional classifiers. However, identification performance has remained to be desired (13). Using a large dataset of Kentucky crash narratives from 2015–2022, we systematically compare three classes of approaches for improving secondary crash data quality: (1) zero-shot LLM prompting; (2) fine-tuned transformer models, and (3) traditional statistical classifiers (representing conventional text-mining techniques). Additionally, we derive practical recommendations on model selection, training-data requirements, incremental processing, and privacy-preserving deployment, thereby translating recent advances in natural language processing into practical implementation for traffic safety practitioners.

The remainder of this paper is organized as follows: Section 2 provides a detailed review of related studies that leverage narrative text mining for crash data quality improvement. Section 3 introduces the classification models evaluated in this study. Section 4 describes the secondary crash dataset, the model calibration process, and the evaluation metrics. Section 5 presents the classification results and analysis. Section 6 discusses practical implications and deployment considerations. Finally, Section 7 concludes the paper.

## 2. EXISTING NARRATIVE MINING STUDIES

Over the decades, the approach to mining crash report narratives has evolved significantly. Early work relied on carefully selected keyword lists and regular expressions(14). For example, Sorock et al. used pre-selected work zone-related words to identify pre-crash vehicle activities and crash types from 6,333 insurance narratives and achieved more accurate results than relying on crash codes (15). Zheng et al. identified secondary crashes via relationship keywords (e.g., ahead, another, earlier) and event keywords (e.g., crash, accident) (16). These methods were easy to follow and interpret and performed well in narrow domains, but they were prone to false positives/negatives and difficult to scale and generalize. For example, phrases like "construction truck" could trigger a false positive if the vehicle was not located in a work zone.

Instead of hand-coding rules, statistical machine learning methods convert unstructured narratives into feature vectors using techniques like term frequency – inverse document frequency (TFIDF) and trains classifiers (e.g. naïve Bayes, logistic regression, support vector machines, and random forest) to automatically identify important features. Tanguy et al. applied support vector machine with linear kernel to classify aviation incident reports based on prelabeled categories (7), while Goh and Ubeynarayana found it best among six classifiers for classifying construction accident types (8). Zhang et al compared four models for secondary crash classification, with logistic regression achieving the highest F1 (0.75) and accuracy (84%) (9). These approaches are easily scalable to large datasets and allow analysts to interpret coefficients to determine which words contributed to the predictions. However, these models relied on bag-of-words features without regard to word order or syntax. As a result, they could miss contextual clues and struggle with negation and causation.

Deep learning addressed some of statistical classifiers' limitations by learning phrase structure and context through convolutional and recurrent layers, using pre-trained word embeddings (dense vector representations of words) such as Word2Vec and GloVe (17, 18). Heidarysafa et al. analyzed railroad accident narratives by combining one-dimensional convolutional layers (to capture local phrase patterns) with Long Short Term Memory/ Gated Recurrent Unit (GRU) recurrent layers (to capture sequential context) and achieved better results compared to traditional machine learning models (19). Sayed et al. experimented with a simple probabilistic Noisy-OR keyword classifier and a GRU recurrent neural network to identify mislabeled or missed work-zone crashes (20). Zou et al. applied similar models to classify Chinese crash narratives by cause (e.g. speed-related vs. turning-related crashes), with text-CNN yielding the best AUC around 0.90 (21). These models captured semantic relationships between terms (e.g., "construction" and "road work" as work-zone related) but required large, labeled data, which may be challenging for certain crash types, such as wrong-way driving or secondary crashes.

Transformer-based models, first introduced in 2018, transformed narrative mining by leveraging attention mechanisms and large-scale pre-training (22). BERT and its variants require comparatively few task-specific examples and can handle long-range dependencies in text (23). Hosseini et al. demonstrated that fine-tuned BERT models outperformed traditional classifiers for wrong-way-driving crashes, achieving an accuracy of 81.6% (24). The analysis showed that BERT could detect clues like "vehicle traveling northbound in southbound lanes" as wrong-way event, despite the complex wording. Oliaee et al. leveraged BERT to analyze over 750 000 crash reports to predict injury severity and showed that the models could be adapted to new jurisdictions with minimal retraining (25). Transformer-based models offer an attractive trade-off: strong language understanding and adaptability with relatively low training effort. However, their fine-tuning demands computing resources and large labeled datasets, which might be challenging for some DOTs.

Most recently, the field has begun exploring Large Language Models (LLMs) such as GPT, Claude, and Llama for crash narrative analysis(11, 12). Transportation researchers are using LLMs for zero-shot classification tasks via prompts and extracting key insights like explanations and event sequences beyond a simple label. Bhagat and Shihab compared GPT-4, LLaMA-2, and Claude to fine-tuned models, and found that LLMs showed strong alignment with human experts in reasoning despite underperforming the fine-tuned BERT variants in accuracy (26). Mumtarin et al. (2023) used ChatGPT, Bard, and GPT-4 for complex queries, such as generating a chronological sequence of events and identifying contributing factors, where traditional models need separate curated pipelines for classification, information extraction, and inference(12). Other studies have used LLMs to uncover under-reported alcohol involvement and to generate pedestrian and bicycle typologies directly from narratives(10, 27). These studies demonstrate that agencies could employ one LLM to handle many narrative analysis tasks, instead of maintaining separate models. However, LLMs are prone to hallucinations when a narrative is ambiguous or lacks detail, leading to misclassification; therefore, careful prompt engineering or fine-tuning is crucial.

## 3.  CLASSIFICATION MODELS UNDER EVALUATION

Crash narratives provide an optimal testbed for various modeling approaches, thanks to their significant variability in length, spelling, and stylistic conventions. Given the frequent mentioning of personally identifying information like names and license numbers by reporting officers, text mining workflows must adhere to stringent privacy regulations. To address this concern, we evaluated three families of text classification models operable solely on local hardware, including foundation-scale large language models, mid-sized transformer encoders, and classical linear models.

Foundation-scale LLMs, including Llama3:70B, DeepSeek-R1:70B, Qwen3:32B, and Gemma3:27B, represent the most advanced open-source systems, pre-trained on extensive web corpora. They enable transportation engineers to extract patterns, causes, and sequential insights from unstructured crash reports via prompt engineering, even amidst spelling or grammatical errors. Llama3:70B, developed by Meta, facilitates expert-level text understanding and generation(28), while DeepSeek-R1:70B from DeepSeek AI excels in logical, step-by-step reasoning, with robust long-context retention(29). Qwen3:32B from Alibaba supports multilingual and extended text processing(30). Gemma3:27B from Google offers a compact, efficient alternative for on-site analysis, balancing performance in summarization and classification tasks on less powerful hardware(31).

Mid-sized transformer encoders, such as BERT(23), DistilBERT(32), RoBERTa(33), Longformer(34), and XLNet(35), occupy an intermediate position, requiring fine-tuning on labeled crash narrative samples (typically thousands of records), after which they operate efficiently offline on workstation-grade GPUs or desktops. BERT from Google provides bidirectional contextual understanding for classifying short descriptions or entity recognition. DistilBERT, a distilled and lighter variant of BERT from Hugging Face, achieves near-equivalent performance with reduced parameters and faster processing. RoBERTa, an enhanced Meta variant, handles linguistic variations robustly for tasks like sentiment analysis in inconsistent reports. Longformer extends this to long documents, enabling timeline extraction from multi-page narratives. XLNet from Google and CMU captures flexible text dependencies for sequencing collision events.

Classical linear models like logistic regression serve as baseline in this study. It relies on manual feature engineering (e.g., word counts) for interpretable classification of incident type or severity on minimal hardware. We performed feature extraction following the same four-step process in (9), involving narrative tokenization, word counting, vectorization, term frequency – inverse document frequency (TF-IDF) normalization.

Table 1 provides more detailed information on these ten models. Model training and testing were conducted on a workstation equipped with 2 AMD EPYC 9454 48-Core Processors with 728G RAM and a NVidia L40S GPU with 48G vRAM.

**Table 1 Classification Models under Evaluation and Their Architecture, Deployment Footprint and Fine-Tuning Approaches**

| Model | Type | Deployment Footprint | Model Architecture | Narrative Mining Capabilities | Fine-Tuning Approach |
|---|---|---|---|---|---|
| Llama3:70B | Large Language Model (LLM) from Meta | 43 GB vRAM GPU | Open-source generative language model, with 70 B parameters and 128 k-token context | excels in zero/few-shot learning; Strong chain-of-thought reasoning; robust to spelling/grammar noise | Prompt engineering / in-context learning |
| DeepSeek-R1:70B | Large Language Model (LLM) from DeepSeek AI | 43 GB vRAM GPU | Distilled to be good at chain-of-thought reasoning; with 70 B parameters; 128 k-token context | High quality step-by-step reasoning; long-context retention | Prompt engineering / in-context learning |
| Qwen3:32B | Large Language Model (LLM) from Alibaba | 20 GB vRAM GPU | Multilingual model; tuned for long context, 32.8 B params; 40 k-token context | Multilingual understanding; efficient long-document handling | Prompt engineering / in-context learning |
| Gemma3:27B | Large Language Model (LLM) from Google | 17 GB vRAM GPU | Google's lightweight multimodal Transformer, 27 B params; 128 k-token context | designed for on-device or edge computing while maintaining strong NLP performance | Prompt engineering / in-context learning |
| BERT (base) | Transformer Encoder from Google | CPU or <2 GB vRAM GPU | Classic "fill-in-the-blank" language model, 110 M params; 512-token context | Strong for sentence-level tasks like classification and NER; bidirectional context captures nuances well | Supervised fine-tuning on labelled crash data |
| DistilBERT (base) | Transformer Encoder from HuggingFace | CPU or <2 GB vRAM GPU | lighter variant of BERT, 66M params; 512-token context | Compact and faster than BERT with near-equivalent performance for classification/NER | Supervised fine-tuning on labelled crash data |
| RoBERTa (base) | Transformer Encoder from Meta | CPU or <2 GB vRAM GPU | Improved BERT by training longer on more text, 125 M params | Better performance on downstream tasks than BERT; robust to variations in text. | Supervised fine-tuning on labelled crash data |
| Longformer (base) | Transformer Encoder from AI2 | 16 GB vRAM GPU | A variant of RoBERTa designed for long documents, 149 M params; 4,096-token context | Efficient for processing extended texts; maintains performance on long-sequence tasks. | Supervised fine-tuning on labelled crash data |
| XLNet | Transformer Encoder from Google/CMU | 4GB vRAM GPU | A generalized autoregressive model that learns bidirectional contexts, 110 M params; 1 k-token context | Captures bidirectional context without masking; strong on tasks requiring order understanding | Supervised fine-tuning on labelled crash data |
| Logistic regression | Traditional Machine Learning | KBs of RAM on CPU | linear classifier that models probabilities; requires manual feature engineering (e.g., TF-IDF vectors) | Transparent coefficients; fast to train and deploy; low computational needs | Supervised fine-tuning on labelled crash data |

# 4. KENTUCKY SECONDARY CRASHES CASE STUDY

## 4.1 Secondary Crash Narratives in 2015-2022

To assess the performance of different classification models, a benchmark dataset of verified secondary crashes was developed using data from the Kentucky State Police Crash Database. This comprehensive database includes crash records from all roadway types across the state, with each record containing a free-text narrative describing how each crash occurs. Although a "Secondary Collision" field has been present on crash reports since 2007, with a help prompt added in 2013 to guide proper use, data accuracy remains a concern. Misinterpretation of the term "secondary collision" has led to frequent confusion between "secondary crashes" (distinct, subsequent incidents) and "secondary collision" (multi-impact events within the same crash). As a result, many crashes are incorrectly marked as secondary (i.e. false positives), while numerous true secondary crashes go unreported (i.e. false negatives).

We undertook manual reviews of crash narratives to confirm whether a crash is secondary or not. Kentucky has around 150 thousand crashes occurring each year; therefore, it is impossible to review every crash narrative. To improve the accuracy and efficiency of secondary crash identification, a spatiotemporal filtering method as detailed in (9) was first used to narrow the review pool to crash pairs occurring in close spatial and temporal proximity. Following the same criteria, we used 2-mile and 100-minute thresholds for access-controlled highways and 0.5-mile and 40-minute thresholds for the rest of the roadways. We also considered crashes occurring in the opposite direction to capture secondary crashes resulting from rubbernecking. We then retrieved the associated crash narratives for these candidate primary and secondary crashes. To streamline review, narratives lacking key indicators – such as "crash," "accident," "incident," "collision," "wreck," codes like "10-46" through "10-49," or crash reference numbers – were excluded. This step eliminated roughly half of the crash pairs. The narratives of the remaining crashes were then manually reviewed by a team of four trained crash data analysts.

Specifically, the crash data in 2015-2022 were analyzed and 16,656 crash narratives were manually reviewed. Of these, 3,803 or 22.8% were confirmed to be true secondary crashes. The annotated dataset was divided chronologically: crash data from 2015 through 2021 were used for model training and calibration, while 2022 data served as a holdout test set to evaluate classification performance. Figure 1 displays the spatial distribution of these crashes, which are mainly concentrated in urban areas and on major highways.

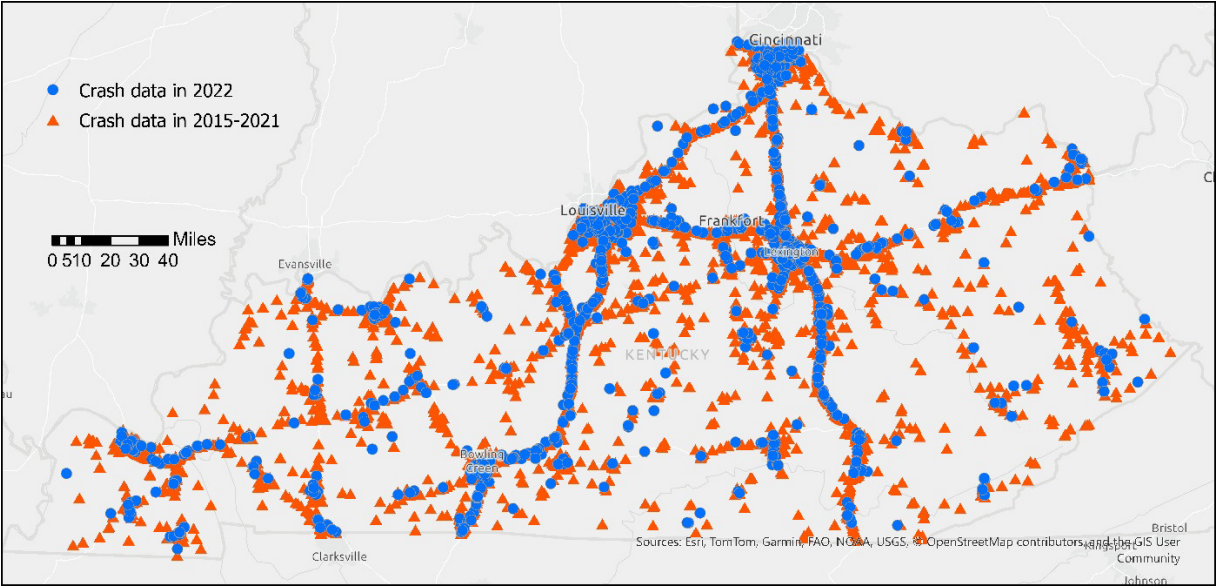

**Figure 1 Spatial distribution of 2015-2022 crashes used for Kentucky secondary crash analysis**

## 4.2 Model Training and Calibration

For open-source LLMs, calibration was performed through structured prompt engineering rather than model parameter fine-tuning. These models operated in a zero-shot or few-shot inference setting, where each narrative was presented alongside a detailed set of secondary crash definition and classification instructions to guide the model's decision-making.

We went through three rounds of prompt refinements based on reviewing classification results from prior round which contain edge cases of false positives and negatives, particularly in ambiguous cases involving emergency vehicles or multi-car chains in severe weather. The insights from the review were incorporated into the prompt, providing clarifying instruction to the model on how to handle confusing cases. The refining process can go on with more iterations, potentially further improving model performance.

The third and final prompt used to feed into model was framed as if the model were a senior traffic safety analyst trained in crash causality assessment. The model was instructed to determine whether a crash was secondary, defined as resulting directly or indirectly from a prior crash, based solely on the information present in the narrative without implying from the model itself. Details of the prompt included:

- A strict definition of secondary crashes, emphasizing causal links such as traffic queues due to a previous crash, debris from earlier crashes, distraction caused by emergency response to a crash, or reactive behavior like abrupt braking or swerving.
- Specific exclusion criteria, clarifying that mere proximity in time or space, or assumptions about traffic conditions, were insufficient without explicit evidence in the narrative.
- Edge-case rules covering weather-induced pileups, wildlife-related crashes, emergency vehicle encounters, and rubbernecking incidents.

- Emphasis on handling ambiguity: if no definitive causal link was identified, the model was to classify the crash as NO, assign a confidence probability near 0.5, and briefly explain the uncertainty.
- The model was instructed to return its decision strictly following a JSON format with three fields:
  {"answer": "<YES / NO>",
   "probability": <float between 0 and 1>,
   "explanation": "< Concise 1-2 sentence explanation referencing narrative details>"}

The five transformer-based models were fine-tuned using labeled training data from 2015–2021 using the HuggingFace Transformers library(36). These models were initialized with pretrained weights and updated through supervised learning with a binary classification head. Cross-validation was used to tune key parameters like learning rates, dropout probabilities, and weight decay and ensure that the models generalized well without overfitting.

For traditional classifiers like Logistic Regression, TF-IDF features were extracted from the same training narratives and hyperparameter tuning was conducted using Scikit-learn library(37)

## 4.3 Classification Performance Metrics

Model performance was evaluated on the 2022 test dataset using standard classification metrics:

- Confusion matrices, i.e., counts of True Positives (TP), True Negatives (TN), False Positives (FP), and False Negatives (FN), were generated for each model to provide insight into their classification behavior. In our context, FP means a crash would be wrongly marked as secondary (over-reporting), whereas FN means a real secondary crash remains unflagged (under-reporting).
- Precision (Positive Predictive Value): for the model-predicted secondary crashes, precision = TP / (TP + FP), i.e. the fraction of predicted "secondary" that are true secondary crashes. This measures how reliable a positive prediction is.
- Recall (Sensitivity): for actual secondary crashes, recall = TP / (TP + FN), i.e. the fraction of true secondary crashes that the model successfully identified. This measures how many of the actual secondary crashes we manage to catch via narrative mining.
- F1 Score: The harmonic mean of precision and recall. It provides a single metric balancing the trade-off between precision and recall. It is useful for comparing models especially when one model might have higher precision and another higher recall. A higher F1 indicates a better balance of catching the most secondary crashes while not flagging too many false ones.
- Accuracy: the overall proportion of crashes correctly classified (secondary or not) out of all cases. Accuracy is straightforward but can be misleading if the classes are imbalanced. In our dataset ~22% of cases were secondary crashes, so accuracy alone might not reflect performance on the secondary crash class; hence it was considered alongside F1.

## 5. CASE STUDY RESULTS AND DISCUSSIONS

Table 2 summarizes the case study results, comparing 10 different modeling approaches for identifying secondary crashes from narrative text. Overall, fine-tuned transformer models slightly outperformed much larger LLMs deployed in a zero-shot inference mode (i.e., without fine-tuning) for this classification task. Fine-tuned RoBERTa was the best-performing model, achieving the highest F1-score at 0.90 with accuracy over 95%. BERT and DistilBERT were only slightly behind (F1 0.88-0.89, accuracy 94%-95%). In contrast, the best large generative models (LLaMA3:70B and DeepSeek-R1:70B, each with tens of billions of parameters) showed strong performance with F1-scores reaching 0.85-0.86, but still marginally below the fine-tuned BERT-family models. The logistic regression baseline lagged far behind (F1 0.66, Accuracy 83%), highlighting the substantial gains from recent more advanced models on this classification task.

Furthermore, the models exhibit notable differences in precision and recall results. RoBERTa and BERT not only achieved the highest overall F1, but also maintained an excellent balance of precision (0.91 and 0.93 respectively) and recall (0.89 and 0.85). This balance indicates they both caught most of the true positive cases while keeping false alarms low. In contrast, XLNet and LongFormer achieved the highest precision (~0.96), meaning they very rarely produced false positives; however, this came at the cost of slightly lower recall (0.80 and 0.77 respectively), so they missed more true secondary crashes than BERT or RoBERTa.

Among LLMs, GEMMA3:27B stood out for its highest recall of all models evaluated (94%), with only 26 false negatives. However, this strength came at the expense of precision (71%), with 170 cases being incorrectly flagged as secondary. In comparison, the larger LLMs (e.g. LLaMA3:70B and DeepSeek-R1:70B) showed balanced precision and recall tradeoff (around 0.85 for both). Meanwhile, Qwen3:32B had the lowest recall (0.76) and lowest F1 (0.79), indicating it struggled the most in identifying positive cases. In summary, the fine-tuned models exhibited the most favorable combination of high precision and recall, whereas the zero-shot LLMs were slightly more error-prone, and the logistic baseline suffered from both low precision and recall.

Another major difference observed is in computational efficiency (see *Run time* in Table 2). Figure 2 illustrates the comparison between model performance and computational efficiency. The y-axis uses the logarithmic scale of the inference time in minutes for processing 1,771 narratives. The LLMs required orders of magnitude more time to process the test set compared to the fine-tuned models. For example, LLaMA3:70B took about 139 minutes, and DeepSeek-R1:70B required over 12 hours to complete the test, reflecting the heavy computational load of prompting large models without task-specific optimization. Qwen3:32B also took a substantial 460 minutes. In contrast, the smaller fine-tuned models have a one-time training cost but then achieve very fast inference. Fine-tuning RoBERTa on 14,885 training narratives took only 13 minutes, after which it classified the 1,771 test narratives in 8 seconds. Other transformer models similarly needed only a few minutes to train and mere seconds to run on the test set. Even the relatively heavy LongFormer model (with longer context handling) took 74 minutes to train and 24 seconds to test, which are still far faster than LLM's inference. The logistic regression baseline was virtually instantaneous but, as noted, its accuracy was much lower.

These results highlight a clear trade-off: zero-shot LLMs eliminate the need for training data, but incur huge runtime costs, whereas fine-tuning smaller transformer models requires labeled data and training time, but yields highly efficient and accurate predictions. In practical terms, if labeled data is available, the fine-tuned models are preferable for deployment given their superior speed and accuracy. Conversely, large LLMs may be useful when training data is scarce or when a single model must handle many tasks.

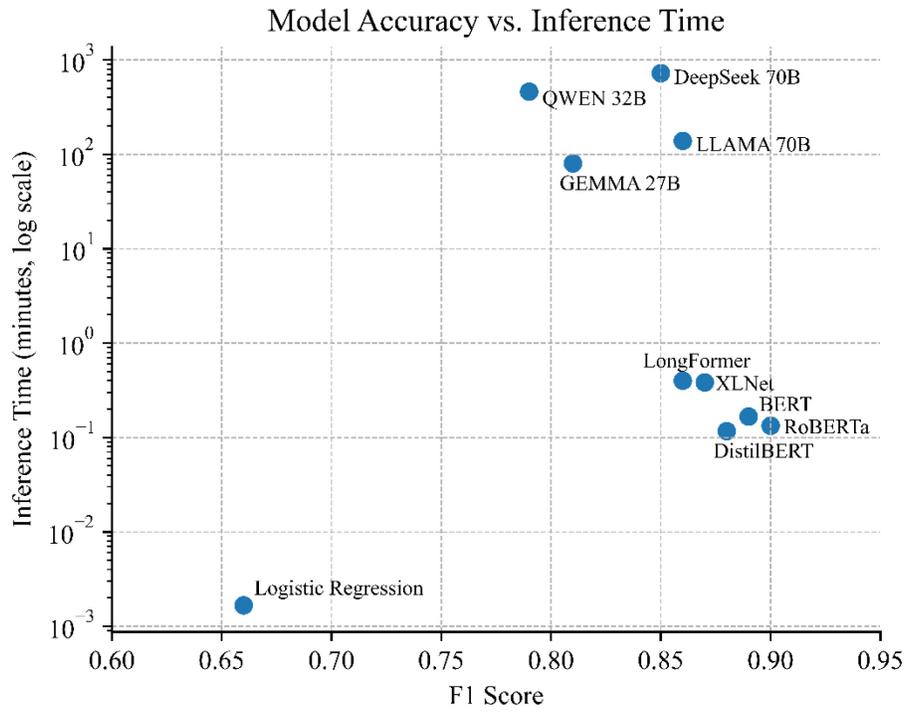

**Figure 2 F1 score vs. inference time for each model, illustrating the trade-off between model performance and computational efficiency**

**Table 2 Classification Performance of 10 models on the 2022 Kentucky Crash Narrative Test Set (N = 1,771)**

| Model | LLAMA 70B | DeepSeek 70B | GEMMA 27B | QWEN 32B | BERT | DistilBERT | XLNet | LongFormer | RoBERTa | Logistic Regression |
|---|---|---|---|---|---|---|---|---|---|---|
| True Negative | 1272 | 1282 | 1164 | 1267 | 1308 | 1276 | 1319 | **1320** | 1296 | 1178 |
| False Positive | 62 | 52 | 170 | 67 | 26 | 58 | 15 | **14** | 38 | 156 |
| False Negative | 62 | 78 | **26** | 105 | 66 | 49 | 89 | 99 | 47 | 145 |
| True Positive | 375 | 359 | **411** | 332 | 371 | 388 | 348 | 338 | 390 | 292 |
| Sum of Falses | 124 | 130 | 196 | 172 | 92 | 107 | 104 | 113 | **85** | 301 |
| F1 | 0.86 | 0.85 | 0.81 | 0.79 | 0.89 | 0.88 | 0.87 | 0.86 | **0.90** | 0.66 |
| Recall | 0.86 | 0.82 | **0.94** | 0.76 | 0.85 | 0.89 | 0.80 | 0.77 | 0.89 | 0.67 |
| Precision | 0.86 | 0.87 | 0.71 | 0.83 | 0.93 | 0.87 | 0.96 | **0.96** | 0.91 | 0.65 |
| Accuracy | 0.93 | 0.93 | 0.89 | 0.90 | 0.95 | 0.94 | 0.94 | 0.94 | **0.95** | 0.83 |
| Run time (1771 narratives as testing set) | no training; test 139 mins | no training; test 723 mins | no training; test 80 mins | no training; test 460 mins | train 13 mins; test 10 secs | train 6.5 mins; test 7 secs | train 38 mins; test 23 secs | train 74 mins; test 24 secs | train 13 mins; test 8 secs | train 4 secs; test 0.1 sec |

Note: Confusion matrix counts (True Negative, False Positive, False Negative, True Positive) are shown, along with total errors ("Sum of Falses"), F1, Recall, Precision, Accuracy, and model run-time on the test set. Pre-trained LLMs (DeepSeek, GEMMA, LLaMA, Qwen) were used in zero-shot inference (no fine-tuning), whereas others were fine-tuned on training data (training time noted). Higher F1, recall, precision, and accuracy values are highlighted in red, as are notably low error counts.

To better understand the impact of model scale, we further examined variant sizes of LLaMA and DeepSeek, as shown in Table 3. The performance of LLaMA drops significantly with smaller size: the 8B-parameter LLaMA model achieved only 85% accuracy (F1 = 0.71), compared to 93% accuracy (F1 = 0.86) for the 70B version. This suggests that the full 70B capacity is needed for LLaMA to perform competitively in zero-shot classification. By contrast, the DeepSeek model maintained strong performance at 32B parameters, which surprisingly slightly exceeded the 70B's recall (0.86 vs 0.82) and obtained a comparable F1 of 0.84 versus 0.85 for the 70B. In other words, DeepSeek 32B performed on par with DeepSeek 70B, despite having less than half the parameters, hinting that one could choose a mid-sized model for a better balance of performance and efficiency. However, the 8B model did not produce usable output (marked "NA" in Table 3), indicating that at very small scales the model likely lacked sufficient capacity to handle complex instructions. In terms of speed, reducing model size unsurprisingly yielded big gains: for instance, LLaMA 8B ran the test in 19 minutes vs 139 minutes for LLaMA 70B, and DeepSeek 32B took 430 minutes vs 723 minutes for 70B. These results reinforce that larger models tend to give better zero-shot accuracy, but smaller models can be more efficient.

**Table 3 Performance and Runtime of Variant LLaMA and DeepSeek Models**

| Model | LLAMA 70B | LLAMA 8B | DeepSeek 70B | DeepSeek 32B | DeepSeek 8B |
|---|---|---|---|---|---|
| True Negative | 1272 | 1185 | 1282 | 1255 | |
| False Positive | 62 | 149 | 52 | 79 | |
| False Negative | 62 | 111 | 78 | 60 | |
| True Positive | 375 | 326 | 359 | 377 | |
| Sum of Falses | 124 | 260 | 130 | 139 | NA |
| F1 | 0.86 | 0.71 | 0.85 | 0.84 | |
| Recall | 0.86 | 0.75 | 0.82 | 0.86 | |
| Precision | 0.86 | 0.69 | 0.87 | 0.83 | |
| Accuracy | 0.93 | 0.85 | 0.93 | 0.92 | |
| Runtime (1771 narratives as testing set) | 139 mins | 19 mins | 723 mins | 430 mins | 33 mins |

Note: DeepSeek 8B model's performance metrics are marked "NA" due to unusable output was returned

## 6. PRACTICAL CONSIDERATIONS FOR DEPLOYMENT

The findings of this study offer practical implications for transportation agencies seeking to enhance crash data quality and analysis workflows. In secondary crash detection, many agencies may consider false negatives more problematic than false positives due to historical underreporting in crash records. False positives, while undesirable, can be rectified through subsequent manual checks. An optimal system should offer high recall to significantly reduce underreporting, while maintaining enough precision to not overwhelm analysts with too many false alarms. In this end, the fine-tuned transformer models like RoBERTa demonstrate this balance and are recommended for deployment, with applicability extending to other crash types such as alcohol involvement, wrong-way driving, or work-zone incidents.

Given different models' strengths as presented in Table 2, one can even adopt an ensemble strategy, combining models (e.g., RoBERTa as primary checker with XLNet and Gemma for parallel analysis). The system can flag records where models disagree for manual review. This strategy can significantly reduce human workload while maintaining high confidence in the automated results. It can also provide a feedback loop to understand model limitations by examining why disagreements occur and improve future models or prompts accordingly.

Meanwhile, zero-shot LLMs, achieving comparable performance to fine-tuned transformer models, offer unparalleled advantages for cross-checking narratives against multiple coded fields (e.g., direction, location, weather, cause) in a single pass without task-specific retraining. This means that agencies can consider just one LLM (such as LLaMa3:70B) as a generalist auditor of crash data, reducing the need to maintain numerous specialized classifiers. It should be mentioned that given the rapid pace of improvement in LLM performance, the efficacy of an LLM-based verification system is likely to increase over time.

Given sensitive personal information in crash narratives, prioritizing privacy is paramount. From this perspective, while online or API-based LLM services relieve the agencies' infrastructure investment burden while also providing most advanced capabilities, they could expose sensitive personal information or violate data-sharing regulations, as highlighted in prior research (11). Therefore, on-premise or local deployment is strongly recommended to keep data entirely under agency control. While the current flagship LLMs are resource-intensive, the trend toward smaller yet capable models and techniques like model quantization will facilitate local deployments. Agencies might initially prototype with smaller open-source models (such as Deepseek-R1: 32B or LLaMA3: 8B) that can run on available hardware, and gradually adopt more powerful models as infrastructure improves.

Another practical consideration is the computational runtime given the large volume of crash records. The fine-tuned transformer models such as RoBERTa and BERT can easily scale to large datasets. For example, about 150,000 Kentucky crash narratives are written each year, and a BERT-like model could classify a full year's worth in minutes of computing time. By contrast, the large LLMs are orders of magnitude slower, therefore it would be impractical to process all these narratives retrospectively. However, this challenge can be mitigated by adopting an incremental workflow: instead of analyzing the yearly database at once, the model could be run on new crash reports as they arrive (or on a weekly/monthly batch). For example, if an agency receives a few hundred new crash records a week, it would take a couple of hours overnight each week to process. The crash data quality verification typically does not require real-time processing; hence, this incremental processing approach is recommended.

Finally, it is crucial to consider the computing resources needed to run these models locally. For many transportation agencies, high-end GPUs represent a non-trivial budgetary decision; therefore, they need to weigh the costs against the benefits. If crash data verification is to be performed continuously and at large volumes (e.g. thousands of reports per week), the time savings and quality gains from an accelerated pipeline could justify the capital expense of dedicated hardware. Moreover, owning the hardware for on-premise deployment addresses the privacy concern noted above. On the other hand, for smaller-scale or less frequent analyses, it might be more cost-effective to use optimized BERT-level models that run on existing computing resources.

In summary, deploying an LLM-assisted crash data verification system is feasible today and stands to become even more practical as technology evolves. Agencies should plan for a solution that takes advantage of LLMs' growing capabilities while mitigating current limitations through thoughtful deployment: keep sensitive data local for privacy, adopt an incremental processing schedule for efficiency, use a mix of models with human oversight to ensure accuracy, and balance hardware constraints and model capability. By considering these factors, practitioners can significantly enhance crash data quality and analytical efficiency in crash reporting, facilitating safer and more informed transportation safety management.

## 7. CONCLUSIONS

This study provides a comprehensive benchmark of narrative-mining algorithms for enhancing crash data quality, focusing on secondary crash identification in Kentucky's 2015-2022 police reports. Through systematic comparison of zero-shot LLMs, fine-tuned transformer models, and traditional classifiers, we demonstrate substantial advancements over conventional methods. Fine-tuned models like RoBERTa and BERT emerged as top performers, achieving F1-scores of 0.90 and 0.89, respectively, with accuracy exceeding 94%. These models effectively balanced precision and recall, minimizing both false positives (over-reporting) and false negatives (under-reporting), which are critical for accurate safety analyses. Zero-shot LLMs, such as LLaMA3:70B (F1: 0.86) and DeepSeek-R1:70B (F1: 0.85), showed competitive results without labeled data but at the expense of prolonged inference times and slightly lower overall performance. The logistic regression baseline lagged significantly (F1: 0.66), underscoring the limitations of bag-of-words approaches in capturing narrative nuances.

Key insights reveal trade-offs in model selection: fine-tuned transformers offer efficiency and accuracy for agencies with modest labeled datasets, while LLMs provide versatility for multi-task verification and exploratory analysis, albeit with higher computational demands. Variant analysis further indicated that mid-sized LLMs (e.g., DeepSeek-R1:32B) can rival larger counterparts in performance while reducing runtime, suggesting opportunities for optimized deployments.

Practically, these findings enable transportation agencies to automate labor-intensive narrative reviews, addressing chronic data quality issues like secondary crash miscoding. By prioritizing recall to combat underreporting, while using ensembles and human oversight for contentious cases, agencies can enhance data accuracy without overwhelming resources. On-premise deployment is recommended to safeguard privacy, especially given sensitive content in crash narratives, with incremental processing mitigating scalability challenges.

In conclusion, this research bridges AI advancements and highway safety practice, offering actionable guidance to reduce errors in crash databases. Improved data quality will support better hotspot identification, countermeasure selection, and resource prioritization, contributing to more effective traffic incident management and safer roadways. Limitations include reliance on Kentucky-specific data and zero-shot LLM inconsistencies due to prompt sensitivity. Cross-jurisdictional validation and advanced prompt engineering could further refine these approaches. As LLM capabilities evolve rapidly, future work should explore hybrid fine-tuning of LLMs, multi-attribute verification (e.g., impairment or work-zone factors), and integration with spatiotemporal methods for holistic crash analysis.

## ACKNOWLEDGEMENT

The authors thank Chris Blackden for crash data extraction, Megan White and Lauren Hayes for narrative review. ChatGPT and Grok were used to proofread the paper's language and correct grammatical errors.

## AUTHOR CONTRIBUTIONS

The authors confirm contribution to the paper as follows: study conception and design: X.Z.; data collection: X.Z.; analysis and interpretation of results: X.Z. and M.C.; draft manuscript preparation: X.Z. All authors reviewed the results and approved the final version of the manuscript.

## DECLARATION OF CONFLICTING INTERESTS

The authors declared no potential conflicts of interest with respect to the research, authorship, and/or publication of this article.

## FUNDING

The research was made possible by funding from Kentucky Transportation Cabinet.